\documentclass[twoside,11pt]{article}

\usepackage{jmlr2e}
\usepackage{float}
\usepackage{natbib}
\usepackage{mathrsfs}
\usepackage{mathtools}
\usepackage{amssymb}
\usepackage{amsmath}
\usepackage{algorithm}
\usepackage{lipsum}
\usepackage[noend]{algpseudocode}
\makeatletter
\def\BState{\State\hskip-\ALG@thistlm}
\makeatother
\DeclareMathOperator*{\argmax}{arg\,max}

\ShortHeadings{Missing Data Imputation for Classification Problems}{Choudhury and Kosorok}
\firstpageno{1}

\begin{document}

\title{Missing Data Imputation for Classification Problems}

\author{\name Arkopal Choudhury \email arkopal@live.unc.edu \\
       \name Michael R.\ Kosorok \email kosorok@bios.unc.edu \\
       \addr Department of Biostatistics\\ 
       University of North Carolina\\
       Chapel Hill, NC 27599-7420, USA}

\editor{Francis Bach, David Blei and Bernhard Scholkopf}

\maketitle

\begin{abstract}
Imputation of missing data is a common application in various classification problems where the feature training matrix has missingness. A widely used solution to this imputation problem is based on the lazy learning technique, $k$-nearest neighbor (kNN) approach. However, most of the previous work on missing data does not take into account the presence of the class label in the classification problem. Also, existing kNN imputation methods use variants of Minkowski distance as a measure of distance, which does not work well with heterogeneous data. In this paper, we propose a novel iterative kNN imputation technique based on class weighted grey distance between the missing datum and all the training data. Grey distance works well in heterogeneous data with missing instances. The distance is weighted by Mutual Information (MI) which is a measure of feature relevance between the features and the class label. This ensures that the imputation of the training data is directed towards improving the classification performance. This class weighted grey kNN imputation algorithm demonstrates improved performance when compared to other kNN imputation algorithms, as well as standard imputation algorithms such as MICE and missForest, in imputation and classification problems. These problems are based on simulated scenarios and UCI datasets with various rates of missingness.
\end{abstract}


\section{Introduction}
Many of the commonly used classification algorithms such as Classification and Regression Trees (CART) \citep{Breiman84CART} and Random Forests \citep{breiman2001random} do not have rigorous techniques for handle missing values in training data. Ignoring the datapoints with missing values and running the classification algorithm on complete cases only leads to loss of vital information \citep{little2002}. The occurrence of missing data is one of the biggest challenges for data scientists solving classification problems in real-world data \citep{duda2012pattern}. These datasets can come from any walk of life, ranging from medical data \citep{troyanskaya2001missing} and survey responses to equipment faults and limitations \citep{le2005estimating}. The reason for missingness can be human error in inputting data, incorrect measurements, non-response to surveys, etc. For example, an industrial database maintained by Honeywell, a company manufacturing and servicing complex equipment, has more than 50\% missing data \citep{lakshminarayan1999imputation} despite regulatory requirements for data collection. In wireless sensor networks, incomplete data is encountered due to sensor faults, local interference or power outage \citep{le2005estimating}. In medical fields, patient health care records are often a by-product of patient care activities rather than an organized research protocol which leads to significant loss of information~\citep{cios2002uniqueness}. This leads to almost every patient record lacking some values as well as each attribute/feature having missing values. More than 40\% of the datasets in the UCI Machine Learning Repository have missing values \citep{newman2008uci}.

Classification problems are aimed at developing a classifier from training data, so that a new test observation can be correctly classified into one of the groups/classes. The class membership is assumed to be known for each observation of the training set whereas the corresponding attributes/features may have some missing values. The test dataset consists of new observations having the corresponding features but no class labels. The goal of the classification problem is to assign class labels to the test set \citep{alpaydin2009introduction}. In our problem setup, we assume that some of the features are missing at random (MAR) for the training as well as the test dataset. One approach to classification is ignoring the observations with missing values and building a classifier. This is only feasible when the missingness is insignificant, however, and it has been demonstrated that even with a 5\% missingness, proper imputation increases the classification accuracy \citep{farhangfar2008impact}.  We focus on imputation of missing values in the training as well as the test dataset so as to improve the overall performance of the classifier on the test data. Our proposed method takes into account the class label during imputation of the training features, and this ensures an overall improvement in classification.

The work related to missing data imputation can be divided into two categories, single imputation and multiple imputation. Single imputation strategies provide a single value for the missing datum. The earliest single value imputation strategy was Mean Imputation \citep{little2002} which ignores the input data distribution by imputing just one value for all missing instances of a feature. Other popular single imputation techniques are hot deck and cold deck imputation~\citep{little2002}, C4.5 \citep{quinlan1993c4} and prediction based models \citep{schafer1997analysis}. C4.5 works well with discrete data but not with numerical data, which has to be discretized to apply the algorithm \citep{tsai2008discretization}. Prediction based models depend on the correct modelling of missing features and the relationship between them. Usually, incomplete datasets obtained from studies cannot be modelled accurately. The problem with single imputation techniques in general is they reduce the variance of the imputed dataset and cannot provide standard errors and confidence intervals for the missing data. They are also very case specific as they can meaningfully impute data only when the model is known or when the data is either numerical or discrete.

To solve the problems of single imputation, multiple imputation strategies generate several imputed datasets from which confidence intervals can be calculated. Multiple imputation is a process where several complete databases are created by imputing different values to reflect the uncertainty about the right values to impute \citep{rubin1977formalizing,farhangfar2007novel}. The earliest multiple imputation technique was the Expectation-Maximization (EM) Algorithm \citep{LR77EM}. The EM Algorithm and its variants such as EM with bootstrapping \citep{honaker2011amelia}, assumes a parametric density function which fails miserably for features without a parametric density. A recent generalization of the EM Algorithm called Pattern Alternating Maximization with Lasso Penalty (MissPALasso) \citep{stadler2014pattern} has been applied to datasets with high dimensionality ($p >> n$), but also assuming normality. Bayesian multiple imputation algorithms have been applied only to multivariate normal samples \citep{li1988imputation, rubin1990efficiently}.

Regression Imputation \citep{gelman2006data} is also a popular multiple imputation technique where each feature is imputed using other features as predictor variables for the regression model. Sequential Regression Multivariate Imputation (SRMI) improves upon this by fitting a sequence of regression models and drawing values from the corresponding predictive distributions \citep{raghunathan2001multivariate}. Incremental Attribute Regression Imputation (IARI) constructs a sequence of regression models to iteratively impute the missing values and also uses the class label of each sample as a predictor variable~\citep{van2016incremental}. In Multiple Imputation using Chained Equations (MICE), the conditional distribution of each missing feature must be specified given the other features \citep{buuren1999flexible}. It is assumed that the feature matrix has a full multivariate distribution from which the conditional distribution of each feature is derived. The full distribution need not be specified, as long as the distribution of each feature is stated, a feature called fully conditional specification \citep{van2007multiple}. MICE can handle mixed types of data. It has options for predictive mean matching, linear regression, binary and polytomous logistic regression, etc., and uses the Gibbs sampler to generate multiple imputations. However, for a given set of conditional distributions, a multivariate distribution may not exist~\citep{van2006fully}. The ideas of MICE and SRMI are combined in the MissForest approach~\citep{stekhoven2011missforest} which fits a random forest on the missing feature, using the other features as covariates and then predicts the missing values. This procedure is iterative and can handle mixed data, complex interactions, and high dimensions.

Machine Learning techniques such as Fuzzy $c$-Means~\citep{sefidian2019missing}, Multilayer Perceptrons (MLP)~\citep{garcia2013classifying} and $k$-Nearest Neighbors (KNN) \citep{batista2002study} are useful non-parametric approaches to imputation of missing data. Various machine learning algorithms such as $k$-Nearest Neighbors (KNN), Support Vector Machines (SVM) and decision trees have been used in imputation by framing the imputation problem as an optimization problem and solving it \citep{bertsimas2017predictive}.  The Nearest Neighbor Imputation (NNI) approach is simple since there is no need to build a predictive model for the data. The basic KNN Imputation (KNNI) algorithm was first used for estimating DNA microarrays with the contribution of the $k$-Nearest Neighbors weighted by Euclidean distance \citep{troyanskaya2001missing}. The sequential KNN method was proposed using cluster-based imputation \citep{kim2004missing}, followed by an iterative variant of the KNN imputation (IKNN) \citep{bras2007improving}, both of which improves on KNNI. The Shelly Neighbors (SN) method improves the KNN rule by selecting only neighbors forming a shell around the missing datum, among the k closest neighbor \citep{zhang2011shell}. The first significant work in improving KNN imputation for classification based problems uses a feature-weighted distance metric based on Mutual Information (MI) as a measure of closeness of a feature to the class label \citep{garcia2009k}. The method is called Mutual Information based $k$-Nearest Neighbor (MI-KNN) Imputation. However, the distance metric used is Euclidean distance, which does not perform well with mixed-type data \citep{huang2006instance}. Alternatively, Grey Relational Analysis is shown to be more appropriate for capturing proximity between two instances with mixed data as well as missingness. Based on this, a Grey KNN (GKNN) imputation approach was built based on Grey distance instead of Euclidean distance and it was shown to outperform traditional KNN imputation techniques \citep{huang2004grey, zhang2012nearest}. This grey distance based KNN imputation is weighted by mutual information between features (measure of inter-feature relevance) and shown to outperform IKNN, GKNN and Fuzzy $k$-Means Imputation (FKMI)~\citep{li2004towards} in most settings, and is called the Feature Weighted Grey $k$-Nearest Neighbor (FWGKNN) method \citep{pan2015missing}. However, this method does not take into account each feature's association with the class label, which is crucial when dealing with classification problems. The FWGKNN method also assumes inter-dependency of features.

We propose a Class-weighted Grey $k$-Nearest Neighbor (CGKNN) imputation method where we calculate the MI of each feature with respect to the class label in the training dataset, use it for calculating the weighted Grey distance between the instances, and then find the $k$-Nearest Neighbors of an instance with missing values. Using $k$-Nearest Neighbors, the missing value is imputed according to the weighted Grey distance. Our contributions can be summarized as follows:
\begin{enumerate}
    \item We use a combination of Mutual Information between each feature and the classifier variable $Y$ to weigh the Grey distance between instances in the feature matrix $X$. This metric is suited for tuning out any unnecessary features for classification and then finding the nearest neighbors relevant for imputation.
    \item We solve an imputation problem with no underlying assumptions on the structure of the feature matrix $X$ except that the data is missing completely at random (MCAR) or missing at random (MAR). Our method (CGKNN) is non-parametric in nature and does not assume any dependence between the individual features. This performs well even when the features are independent of each other.
    \item The proposed CGKNN imputation method is suited well for classification problems where the training as well as the test datasets have missing values. The feature matrix can be mixed-type, i.e., have categorical and numeric data. Our method is suitable for mixed-data classification problems faced with missing values. Moreover, our problem approach takes much less time to initialize than the most similar alternative method, Feature Weighted Grey $k$-Nearest Neighbor (FWGKNN).
\end{enumerate}

The remainder of this paper is organized as follows. In Section 2, we review the KNN imputation techniques used in previous work and then provide a detailed outline of our method. We also discuss how it can be extended to the test dataset for classification and also derive the time complexity of our algorithm. In section 3, we test our proposed method against 6 standard methods in simulation settings. We evaluate our imputation method (CGKNN) in different simulation settings with classification where we artificially introduce missingness. We compare it with standard multiple imputation algorithms MICE and MissForest as well as the previous KNN based algorithms, Iterative KNNI (IKNN), Mutual Information based KNNI (MI-KNN), Grey KNNI (GKNN) and Feature-Weighted Grey KNNI (FWGKNN). In section 4, we demonstrate how our algorithm performs with classification tasks involving 3 UCI Machine Learning Repository datasets. We also check for improvement of classification accuracy after imputation of the missing data. Our method gives the best classification performance out of all evaluated methods. We conclude with a discussion and scope for future work in section 5.

\section{Methodology}
In this section, we pose the missing data problem which is encountered in classification tasks. We introduce the nearest neighbor (NN) approach and the previous works done on implementing variations on the KNN imputation approach. This is followed by the concepts of mutual information (MI) and grey relational analysis (GRA) used by our method of Class-weighted Grey $k$-Nearest Neighbor (CGKNN) imputation approach. We then formalize our imputation algorithm and calculate its time complexity.

\subsection{Formulation of the Problem}
Let $X = \{X_{i}\}_{i=1}^n$ be an $n \times p$-dimensional dataset of $n$ independent observations with $p$ features/attributes and $Y$ a response variable of the class labels influenced by $X$. We assume no dependence structure between the features in $X$. Let $D$ be an $n \times p$-dimensional matrix indicating the missingness of corresponding values in the dataset $X$. In practice, we obtain a random sample of size $n$ of incomplete data associated with a population $(X, Y, D)$, called the training data~\citep{ELS09} used to train the classifier
\begin{equation}\label{setup}
    \mathscr{D} = \{(X_i, Y_i, D_i)\}_{i=1}^n,
\end{equation}
where all the class labels in $\{Y_i\}_{i=1}^n$ are observed, $X_i = (X_{ij})_{j=1}^p = (X_{i1},...,X_{ip})$ represents the $p$ features of the $i$-th observation along with indicator variables $D_i = (D_{ij})_{j=1}^p $ such that
\begin{equation}\label{eq:d}
D_{ij} = \begin{cases}
             0, & \text{$X_{ij}$ is missing}\\
             1, & \text{otherwise}.
         \end{cases}
\end{equation}
Without loss of generality, we can assume for each $i$, the observation $X_i = (X_{ij})_{j=1}^p$ contains $p_0$ categorical features for $j \in \{1,2,...,p_0\}$ and $p_1$ continuous features for $j \in \{p_0+1,...,p_0+p_1\}$ such that $p_0 + p_1 = p$. Let the $j$-th categorical feature contain $k_j$ different values and the $j$-th continuous variable representing the $(p_0 + j)$-th feature of $X_i$, indexed by $j\in\{1,...,p_1\}$ take values from a continuous set $C_j \subset \mathbb{R}$. For each of the categorical features, we can map the $k_j$ different values to the first $k_j$ natural numbers, such that $X \in \{1,...,k_1\} \times ... \times \{1,...,k_{p_0}\} \times C_1 \times ... \times C_{p_1} \subset \mathbb{R}^p$.

In this setting, we can assume that $\{(X_i, Y_i)\}_{i=1}^n$ satisfy the model
\begin{equation}\label{eq:class}
Y_i = g(X_i), \quad i=1, 2,..., n,
\end{equation}
where $g(.)$ is an unknown function mapping a $p$-dimensional number (belonging to a subspace of $\mathbb{R}^p$) to a discrete set $\mathscr{G}$ representing the class labels and $Y_i \in \mathscr{G}$. We assume that $\mathscr{G}$ contains $m$ values and thus the classification problem is based on $m$ classes.

The task of any classification algorithm is to use the training dataset $\{(X_i,Y_i)\}_{i=1}^n$ to estimate $g(.)$, which is referred to as `training' a classifier $\hat{g}(.)$. Given a new set of $\ell$ observations, $X' = \{X'_i\}_{i=1}^\ell$, called the test dataset~\citep{ELS09}, the classifier predicts the corresponding class $Y' = \{Y'_i\}_{i=1}^\ell$ using $\hat{Y_i'} = \hat{g}(X_i').$ Note that the test dataset $X'$ can also contain missing values. Many classification algorithms have been shown to perform better in terms of classification accuracy after imputing the missing values in the feature matrix $X$~\citep{farhangfar2008impact, luengo2012choice} and then training the classifier. In this paper, we propose a nearest neighbor based imputation algorithm which is used to impute the missing values in X and then train the classifier $\hat{g}(.)$. The same algorithm can be extended to the test dataset and impute the missing values in $X'$.

In general, there are three different missing data mechanisms as defined in the statistical literature \citep{little2002}:
\begin{enumerate}
    \item Missing Completely at Random (MCAR): When the missingness of $X$ does not depend on the missing or observed values of $X$. In other words, using $D$ is as defined in \eqref{eq:d},     \begin{equation} \label{4}
            P(D|X) = P(D), \quad \text{for all}\; X
        \end{equation}
    \item Missing at Random (MAR): When the missingness of $X$ depends on the observed values of $X$ but not on the missing values of $X$. If we split the training dataset $X$ into two parts, observed $X_{obs}$ and missing $X_{mis}$, then
        \begin{equation} \label{5}
            P(D|X) = P(D|X_{obs}), \quad \text{for all}\; X_{mis}
        \end{equation}
    \item  Not Missing at Random (NMAR): When the data is neither MCAR or MAR, the missingness of $X$ depends on the missing values of $X$ itself. This sort of missingness is difficult to model as the observed values of X give biased estimates of the missing values.
        \begin{equation} \label{6}
            P(D|X) = P(D|X_{obs},X_{mis})
        \end{equation}
\end{enumerate}
For our problem, we assume that the missing data mechanism of $X$ is either MCAR or MAR. 

\subsection{k-Nearest Neighbors (KNN) Imputation Algorithm}
KNN is a lazy, instance-based learning algorithm and is one of the top 10 data mining algorithms \citep{wu2008top}. Instance-based learning is based on the principle that instances within a dataset will generally exist in close proximity with other cases that have similar properties \citep{aha1991instance}. The KNN approach has been extended to imputation of missing data in various datasets \citep{troyanskaya2001missing}. In general, the KNN imputation is an appropriate choice when we have no prior knowledge about the distribution of the data. Given an incomplete instance, this method selects its $k$ closest neighbours according to a distance metric, and estimates missing data with the corresponding mean or mode. The mean rule is used to predict missing numerical features and the mode rule is used to predict missing categorical features. KNN imputation does not create explicit predictive models, because the training dataset is used as a lazy model. Also, this method can easily treat cases with multiple missing values.

\subsubsection{Distance Metric for Mixed Data} \label{heom}
Let there be two input vectors, $X_a$ and $X_b$ - whose features can be both continuous as well as categorical. The Heterogeneous Euclidean Overlap Metric (HEOM)~\citep{batista2003analysis}, denoted as $d(X_a,X_b)$, is defined as
\begin{eqnarray}
    d(X_a,X_b) &=& \sqrt{\sum_{j=1}^p d_j(X_{aj},X_{bj})^2}\;, \label{7} \\
    d_j(X_{aj},X_{bj}) &=& \begin{cases}
                            1, & \text{$D_{aj}*D_{bj} = 0$ from \eqref{eq:d}}\\
                            d_0(X_{aj},X_{bj}), & \text{$X_j$ is categorical}\\
                            d_N(X_{aj},X_{bj}), & \text{$X_j$ is quantitative}
                         \end{cases} \label{8} \\
    d_0(X_{aj},X_{bj}) &=& \begin{cases}
                            0, & X_{aj} = X_{bj} \\
                            1, & X_{aj} \neq X_{bj}
                         \end{cases} \label{9} \\
    d_N(X_{aj},X_{bj}) &=& \frac{|X_{aj}-X_{bj}|}{max(X_{.j})-min(X_{.j})}\;, \label{10}
\end{eqnarray}
where $max(X_{.j})$ means the maximum value of $n$ observations of feature $X_{.j}$ and $min(X_{.j})$ means the minimum value of  $X_{.j}$ when it is quantitative. The distance ranges from $0$ to $1$ and also takes the value 1, when either of the observations are missing.

However, to effectively apply the KNN imputation approach, a challenging issue is the optimal value of $k$, and the other is selecting neighbours. The optimal $k$-value can be selected using only non-missing parts \citep{kim2004missing}. This $k$-value estimating procedure considers some elements of the non-missing parts as artificial missing values, and finds an expected $k$-value that produces the best estimation ability for the artificial missing values. This method does not perform well when there are large amounts of missing data. In the proposed approach we determine this parameter optimally using cross validation \citep{stone1974cross}.

\subsubsection{KNN Imputation} \label{knn}
Suppose the $j$-th input feature of $X_i$ is missing (i.e., $D_{ij} = 0$ from \eqref{eq:d}) and has to be imputed. After the distances (calculated by the HEOM defined in \eqref{heom}) from $X_i$ to all other training instances ($\{X_k\}_{k=1,k \neq i}^n$) are computed, its $k$-nearest neighbours are chosen from the training set. In our notation, $\mathscr{V}_{X_i} = \{v_\ell\}_{\ell=1}^k$ represents the set of $k$-nearest neighbors of $X_i$ arranged in increasing order of its distance as defined by~\eqref{7}-\eqref{10}. The $k$-closest cases are selected after instances with missing entries in the incomplete feature are imputed using mean or mode imputation, depending on the type of feature \citep{troyanskaya2001missing}. Once its $k$-nearest neighbours have been chosen, the unknown value is imputed by an estimate from the $j$-th feature values of $\mathscr{V}_{X_i}$.

For continuous variables, the imputed value ($\widetilde{X}_{ij}$) is $\widetilde{X}_{ij} = (1/k) \sum_{\ell=1}^k v_{\ell j}$. One obvious refinement is to weight the contribution of each $v_\ell$ according to their distance to $X_i$~\citep{dudani1976distance}, such that
\begin{equation}\label{11}
    \widetilde{X}_{ij} = \frac{1}{kW} \sum_{\ell=1}^k w_\ell\;v_{\ell j}, \quad w_\ell = \frac{1}{{d(X_i,v_\ell)}^2}, \quad W = \sum_{\ell=1}^k w_\ell,
\end{equation}
where $w_\ell$ denotes the corresponding weight of the $\ell$-th nearest neighbour $v_\ell$ and $d(X_i,v_\ell)$ is as defined in \eqref{7}-\eqref{10}.

For categorical or discrete variables, we choose among the discrete values of $X_{.j}$. using the values of the $j$-th input features in $\mathscr{V}_{X_i}$. A popular choice is to impute the mode of $\{v_\ell\}_{\ell=1}^k$ to $\widetilde{X}_{ij}$, where all neighbours have the same importance in the imputation stage \citep{troyanskaya2001missing}. An improvement to this is assigning a weight $\alpha_\ell$ to each $v_\ell$, with closer neighbours having greater weights. The category of $\widetilde{X}_{ij}$ is chosen by the category whose weights sum up to the highest value in $\mathscr{V}_{X_i}$. Using an approach similar to a distance-weighted KNN classifier \citep{dudani1976distance}, a suitable choice of $\alpha_\ell$ is
\begin{equation}\label{12}
    \alpha_\ell(X_i) = \frac{d(v_k,X_i) - d(v_\ell,X_i)}{d(v_k,X_i) - d(v_1,X_i)}\;,
\end{equation}
where $d(.,.)$ is defined in \eqref{7}-\eqref{10} and $\alpha_\ell$ is assigned a value of 1 when $d(v_k,X_i) = d(v_1,X_i)$, that is, all the distances are equal. Otherwise, for $k\;(>1)$ neighbors, $0 \leq \alpha_\ell \leq 1$.

Suppose the $j$-th input feature $X_{.j}$ has $S$ possible discrete values with $K_s$ be the number of samples from $\mathscr{V}_{X_i}$ that belong to category $s,\; s = 1,2,...,S$. Now for each possible category, $\alpha_{X_i}^s$ is calculated by
\begin{equation}\label{13}
    \alpha_{X_i}^s = \sum_{\ell=1}^{K_s} \alpha_\ell (X_i).
\end{equation}
Then, the category $s^*$ imputed to ${\widetilde{X}_{ij}}$ is
\begin{equation}\label{14}
    s^* = \argmax_s \{\alpha_{X_i}^s\}.
\end{equation}

\begin{algorithm}[H]
\caption{Iterative KNN (IKNN) Imputation \citep{garcia2009k}}
\label{iknni}
\begin{algorithmic}
\State \textbf{Input:} $(X,Y,D)$ with $X \subset \mathbb{R}^{n\times p}$ containing missing entries and $Y$ the class labels.
\State \textbf{Output:} Imputed feature matrix $\widetilde{X}$ with no missing values.
\State \textbf{Procedure:}
\begin{enumerate}
    \item \textbf{Initialization:} Given the training dataset $X$, the missing values of $p_0$ categorical features are imputed by mode imputation and the missing values of the $p_1$ continuous features are imputed by mean imputation using the observed data. We call the initially imputed matrix $\widetilde{X}^0$.
    \item \textbf{Choosing $k$:} We use this imputed matrix, $\widetilde{X}^0$, to calculate the optimum value of $k$ using 10-fold cross validation \citep{stone1974cross} to minimize the misclassification rate of predicting the class labels $Y$. This is the $k$ used for choosing the nearest neighbors.
    \item \textbf{Iterative Step:} Consider the iteration number $t\;(\geq 1)$. In the $t$-th iteration, the imputed matrix $\widetilde{X}^t$ is obtained by imputing the missing continuous features (with corresponding $D_{ij} = 0$) using \eqref{11} and missing categorical features using \eqref{12}-\eqref{14}. This step is repeated until the stopping criteria is reached.
    \item \textbf{Stopping Criterion:} We stop at the $d$-th iteration when a stopping criteria is met. The stopping criteria we propose is
    \begin{equation} \label{15}
        \max_{i,j: D_{ij}=0} |\widetilde{X}^d_{ij} - \widetilde{X}^{d-1}_{ij}| < \epsilon,
    \end{equation}
    where $\epsilon = 10^{-4}$ is the chosen accuracy level.
\end{enumerate}
\end{algorithmic}

\end{algorithm}

\subsection{Mutual Information (MI) for Classification} \label{MInotion}
We can see that the above imputation algorithm does not consider the class label $Y$ while computing the $k$-nearest neighbors. We can solve this using an effective procedure where the neighbourhood is selected by considering the input feature relevance for classification~\citep{garcia2009k}. This input feature relevance  for classification is measured by calculating the Mutual Information (MI) between the feature $X_{.j}$ and the class variable $Y$.

\subsubsection{Notion of MI}
The entropy $H(X)$ of a random variable, $X$, measures the uncertainty of the variables. If a discrete random variable $X$ has $\mathscr{X}$ alphabets and the probability density function (pdf) is $p(x) = Pr\{X = x\},\;\; x \in \mathscr{X}$, then the entropy is defined by
\begin{equation}\label{16}
    H(X) = -\sum_{x \in \mathscr{X}} p(x)\;log\; p(x).
\end{equation}
Here the unit of entropy is the bit and the base of the logarithm is 2. The joint entropy of $X$ and $Y$ is defined as
\begin{equation}\label{17}
    H(X,Y) = -\sum_{x \in \mathscr{X}} \sum_{y \in \mathscr{Y}} p(x,y)\;log\; p(x,y),
\end{equation}
where $p(x,y)$ is the joint pdf of $X$ and $Y$, both of them being discrete.

The conditional entropy quantifies the resulting uncertainty of $X$ given $Y$, given by
\begin{equation}\label{18}
    H(Y|X) = -\sum_{x \in \mathscr{X}} \sum_{y \in \mathscr{Y}} p(x,y)\;log\; p(y|x),
\end{equation}
where $p(y|x)$ is the conditional pdf of $Y$ given $X$. The mathematical definition of MI quantifying the dependency of two random variables is given by
\begin{equation}\label{19}
    I(X;Y) = \sum_{x \in \mathscr{X}} \sum_{y \in \mathscr{Y}} p(x,y)\;log\; \frac{p(x,y)}{p(x)p(y)}\;.
\end{equation}
For continuous random variables, the definitions of
entropy and of MI are
\begin{eqnarray}
    H(X) &=& -\int_{x} p(x)\;log\; p(x)\; dx \label{20},\\
    H(Y|X) &=& -\int_x \int_y p(x,y)\;log\; p(y|x)\; dx \label{21},\\
    I(X;Y) &=& \int_{x} \int_{y} p(x,y)\;log\; \frac{p(x,y)}{p(x)p(y)}\;dx\;dy \label{22}.
\end{eqnarray}
The entropy and MI satisfy the following relationship
\begin{equation}\label{23}
    I(X;Y) = H(Y) - H(Y|X),
\end{equation}
which is the reduction of the uncertainty of $Y$ when $X$ is known~\citep{kullback1997information}. The MI can also be rewritten as $I(X;Y) = H(X) + H(Y) - H(X,Y)$, where $H(X,Y)$ is the joint entropy of $X$ and $Y$. When the variables $X$ and $Y$ are independent, the MI of the two variables is zero. Compared to the Pearson correlation coefficient which only measures linear relationships, MI can measure any relationship between variables \citep{kullback1997information}.

\subsubsection{Computation of MI in Classification Problems}
Consider the class label $Y$ for an $m$-class classification problem and let the number of observations in the $y$-th class be $n_y$ such that $n_1+n_2+...+n_m = n$, as mentioned in \eqref{setup}. In terms of classification problems, we are interested in finding the relevance of the $j$-th feature $X_{.j}$ with the class label $Y$, which is measured by their Mutual Information (MI)
\begin{equation} \label{24}
    I(X_{.j};Y) =  H(Y) - H(Y|X_{.j}),
\end{equation}
In this equation, the entropy of class variable $Y$ can be computed using \eqref{16} as
\begin{equation} \label{25}
    \Hat{H}(Y) = -\sum_{y=1}^{m} \Hat{p}(y)\;log\; \Hat{p}(y).
\end{equation}
We can easily estimate $p(y)$ by $\Hat{p}(y)=n_y/n$. The estimation of $H(Y|X_{.j})$ is given by \eqref{18} when $X_{.j}$ is discrete and by \eqref{21} when $X_{.j}$ is continuous. For discrete feature variables, estimating the probability densities is straightforward by means of a histogram approximation \citep{kwak2002input}.

For continuous features, entropy estimation is not straightforward due to the problem of estimation of $p(y|x_{.j})$, where $y$ is discrete and $x_{.j}$ is continuous. Note that we need to estimate the conditional density of $x_{.j}$ at the $m$ classes represented by $y$ and not the joint density. We can use a Parzen window estimation approach to estimate $p(x_{.j})$ \citep{kwak2002input} given by
\begin{equation} \label{26}
    \Hat{p}(x_{.j}) = \frac{1}{n}\sum_{i=1}^{n} \phi(x_{.j} - x_{ij},h),
\end{equation}
where $\phi(.)$ is the window function and $h$ is smoothing parameter. Rectangular and Gaussian functions are suitable window functions~\citep{duda2012pattern} and if $h$ is selected appropriately, $\Hat{p}(x_{.j})$ converges to $p(x_{.j})$ \citep{kwak2002input}. We can calculate $p(x_{.j}|y)$ using the Parzen window approach
\begin{equation} \label{27}
    \Hat{p}(x_{.j}|y) = \frac{1}{n_y}\sum_{i\in I_y} \phi(x_{.j} - x_{ij},h),
\end{equation}
where $I_y$ is the set of observations with class label $y$. Finally, we the use Bayes rule and \eqref{26}-\eqref{27} to estimate $p(y|x_{.j})$ as
\begin{equation} \label{28}
    \Hat{p}(y|x_{.j}) = \frac{\Hat{p}(x_{.j}|y)\;\Hat{p}(y)}{\Hat{p}(x_{.j})}\;,
\end{equation}
and then estimate $H(Y|X_{.j})$ from \eqref{21} by replacing the integral by summation over training observations and using $p(x,y) = p(x)\;p(y|x)$ to arrive at
\begin{equation} \label{29}
    \widehat{H}(Y|X_{.j}) = -\sum_{x_{.j}}\Hat{p}(x_{.j}) \sum_{y=1}^{n_y} \Hat{p}(y|x_{.j}) \;log\;\Hat{p}(y|x_{.j}).
\end{equation}
Using the Parzen window approach, along with \eqref{25} and \eqref{29}, we can calculate the Mutual Information from \eqref{24} between any feature $X_{.j}$ and the class variable $Y$, which measures the relevance of the feature $X_{.j}$ in classification. Using this, a weight $\lambda_j$ is assigned to each feature $X_{.j}$ in Mutual Information based KNNI (MI-KNN) \citep{garcia2009k}, such that
\begin{equation} \label{30}
    \lambda_j = \frac{I(X_{.j};Y)}{\sum_{j'=1}^{p}{I(X_{.j'};Y)}} \;,
\end{equation}
and then the distance between instances is calculated similar to \eqref{7}:
\begin{equation} \label{31}
     d_I (X_a,X_b) = \sqrt{\sum_{j=1}^p \lambda_j d_j(X_{aj},X_{bj})^2}, 
\end{equation}
where $d_j(X_{aj},X_{bj})$ is as defined in \eqref{8}. Using this feature relevance weighted distance, replacing $d$ with $d_I$ (from \eqref{31}) in \eqref{11}-\eqref{12}, and following Algorithm \ref{iknni}, we obtain the MI-KNN imputation algorithm~\citep{garcia2009k}.

\subsection{Grey Relational Analysis (GRA) based KNNI}
Grey System Theory (GST) has been developed to tackle systems with partially known and partially missing information \citep{ju1982control}. The system was named grey since missing data is represented by black whereas known data is white, and this system contains both missing and known data. To obtain Grey-based $k$-nearest neighbors, we used Grey Relational Analysis (GRA) in our algorithm which is calculating Grey Distance between two instances. Grey distance measures similarity of two random instances, which involves the Grey Relational Coefficient (GRC) and the Grey Relational Grade (GRG).

Consider the setup in \eqref{setup} where the training dataset has $n$ observations and $p$ features. The Grey Relational Coefficient (GRC) between two instances/observation $X_a$ and $X_b$, when the $j$-th feature is continuous and observed for both instances, is
\begin{equation} \label{32}
    GRC\;(X_{aj},X_{bj}) = \frac{\Delta_{\min}\; + \;\rho\;\Delta_{\max}}{|X_{aj} - X_{bj}|\; + \;\rho\;\Delta_{\max}}\;,
\end{equation}
where $\Delta_{\min} = \min_c \min_k |X_{ak} - X_{ck}|$, $\Delta_{\max} = \max_c \max_k |X_{ak} - X_{ck}|$, $\rho \in [0,1]$ (usually $\rho = 0.5$ is taken \citep{ju1982control}), $b, c \in \{1,2,...,n\}$, and, $k, j \in \{1,2,...,p\}$ and for categorical feature $j$, $GRC(X_{aj},X_{bj})$ is 1 if they have the same values, 0 otherwise. If either $X_{aj}$ or $X_{bj}$ is missing, then $GRC(X_{aj},X_{bj})$ is 0. The Grey Relational Gradient (GRG) between the instances is defined as: 
\begin{equation} \label{33}
    GRG (X_a,X_b) = \frac{1}{p} \sum _{j=1}^p GRC(X_{aj},X_{bj}),
\end{equation}
where $a \in \{1,2,...,n\}$. We note that if $GRG(X_a,X_b)$ is larger than $GRG(X_a,X_c)$ then the difference between $X_a$ and $X_b$ is less than the difference between $X_a$ and $X_c$, which is the opposite of the Heterogenous Euclidean Overlap Metric (HEOM) \eqref{7} defined in Section \ref{heom}. The Grey Relational Gradient satisfies the following axioms which makes it a distance metric \citep{ju1982control}:
\begin{enumerate}
    \item Normality: The value of $GRG(X_a,X_b)$ is between 0 and 1.
    \item Dual Symmetry: Given only two observations $X_a$ and $X_b$ in the relational space, then $GRG(X_a,X_b) = GRG(X_b,X_a)$.
    \item Wholeness: If 3 or more observations are made in the relational space then $GRG(X_a,X_b)$ is generally not equal to $GRG(X_b,X_a)$ for any $b$.
    \item Approachability: $GRG(X_a,X_b)$ decreases as the difference between $X_{aj}$ and $X_{bj}$ increases, other values in \eqref{32} and \eqref{33} remaining constant. 
    
\end{enumerate}

GRA is generally preferred over metrics such as Heterogeneous Euclidean Overlap Metric (HEOM) for grey systems with missing data \citep{huang2004grey}. It gives us a normalized measuring function for both missing/available and categorical/continuous data due to its normality. It also gives whole relational orders due to its wholeness over the entire relational space. So instead of $d(X_a,X_b)$ in \eqref{7}, if we use $GRG(X_a,X_b)$ to select the $k$-nearest neighbors and then proceed with the KNN Imputation technique without using weights, then it becomes Grey KNN (GKNN) Imputation \citep{zhang2012nearest}.

\subsection{Transformation of the Data} \label{trans}
Before we apply our version of the algorithm, we make some transformation of the continuous features contained in the training dataset, since we deal with a wide variety of features whose ranges vary vastly. For example, the range of marks in a 10 point exam would be less than the range of marks for a 100 point exam, and both these marks may be in the same training dataset. The distance metric and subsequently the $k$-nearest neighbor would be biased unless the ranges of the continuous variables are normalized. In our algorithm, we transformed the $j$-th feature of observation $X_i$ as
\begin{equation} \label{34}
    X'_{ij} = \frac{\max_a X_{aj} - X_{ij}}{\max_a X_{aj} - \min_a X_{aj}} \;,
\end{equation}
where $a,i \in \{1,2,...,n\}$ and $j \in \{1,2,...,p\}$. Thus \eqref{34} ensures all the continous variables are between 0 and 1. Note that the distance metric associated with categorical variables (Euclidean or Grey-based) lie within 0 and 1 as well.

\subsection{The Proposed Class-weighted Grey k-Nearest Neighbor (CGKNN) Algorithm}
We consider the class weight $\lambda_j$ associated with the $j$-th attribute $X_{.j}$ and use this to weigh the Grey Relational Gradient (GRG) between two observations $X_a$ and $X_b$ as follows
\begin{equation} \label{35}
    GRG(X_a,X_b) = \sum_{j=1}^p \lambda_j\; GRC(X_{aj},X_{bj}).
\end{equation}
Since $GRG(X_a,X_b)$ increases for closer neighbors unlike the other distance metrices, we use $d(X_a,X_b) = 1 - GRG(X_a,X_b)$ in section \ref{knn} and then measure the distance between instances to choose the $k$-nearest neighbors, $\{v_\ell\}_{\ell = 1}^k$. From \eqref{11}, we derive that the corresponding weights of $v_\ell$ are
\begin{equation} \label{36}
    w_\ell = \frac{1}{(1 - GRG(X_i,v_\ell))^2}.
\end{equation}
Using these weights, we impute the continuous variables, and the new definition of $d(X_a,X_b)$ in \eqref{12}-\eqref{14} is used to impute the categorical variables for our Class-weighted Grey KNN (CGKNN) Imputation Algorithm.

The overall framework for our proposed algorithm is formalized as Algorithm \ref{cgknn} below.
\begin{algorithm}[H]
\caption{Class-weighted Grey $k$-Nearest Neighbor (CGKNN) Imputation}
\label{cgknn}
\begin{algorithmic}
\State \textbf{Input:} $(X,Y,D)$ with $X \subset \mathbb{R}^{n\times p}$ containing missing entries and $Y$ the $m$ class labels.
\State \textbf{Output:} Imputed feature matrix $\widetilde{X}$ with no missing values.
\State \textbf{Procedure:}
\begin{enumerate}
    \item \textbf{Data pre-processing:} First we transform the continuous features of $X$ as suggested by \ref{trans} using \eqref{34} so that their ranges equal 1.
    
    \item \textbf{Initialization:} We use the class labels in $Y$ to split $X$ into $\{X^y\}_{y=1}^m$. For each class $y$, given $X^y$, we pre-impute the missing values of $p_0$ categorical features by mode imputation and the missing values of the $p_1$ continuous features by mean imputation using the observed data in that class. We call the initially imputed matrix $\widetilde{X}^{y,0}$. Repeat this for $y = \{1,2,...,m\}$. Fuse them to form $\widetilde{X}^{0} = \{\widetilde{X}^{y,0}\}_{y=1}^m$.
    
    \item \textbf{Mutual Information:} Calculate the mutual information or the class weights $\lambda_j$ of the attributes $X_{.j}$ using \eqref{24}-\eqref{30}.
    
    \item \textbf{Choosing $k$:} We use this imputed matrix, $\widetilde{X}^0$, to calculate the optimum value of $k$ using 10-fold cross validation~\citep{stone1974cross} to minimize the misclassification rate of predicting the class labels $Y$. This is the $k$ used for choosing the nearest neighbors.
    
    \item \textbf{Iterative Step:} Consider the iteration number $t\;(\geq 1)$ and class number $y$. For each instance $i$ in the class $y$ which has a missing value, calculate the GRG of that instance with all other instances of the class $y$. After sorting the GRG in descending order, the first $k$ observations are chosen to form the set of nearest neighbors $\{v_\ell\}_{\ell = 1}^k$. Using the weights $w_\ell$ as described in \eqref{36}, the imputed matrix $\widetilde{X}^{y,t}$ is obtained by imputing the missing continuous features (with corresponding $D_{ij} = 0$) using the steps in \eqref{11} and the missing categorical features using~\eqref{12}-\eqref{14} with $d(X_a,X_b) = 1 - GRG(X_a,X_b)$. This is repeated for each $y$ until all missing values are imputed to obtain $\widetilde{X}^{t} = \{\widetilde{X}^{y,t}\}_{y=1}^m$. If the stopping criterion is not met, then the iteration on $t$ continues.
    
    \item \textbf{Stopping Criterion:} We stop at the $d$-th iteration when a stopping criteria is met. The stopping criteria we propose is
    \begin{equation} \label{15}
        \max_{i,j: D_{ij}=0} |\widetilde{X}^d_{ij} - \widetilde{X}^{d-1}_{ij}| < \epsilon,
    \end{equation}
    where $\epsilon = 10^{-4}$.
\end{enumerate}
\end{algorithmic}
\end{algorithm}

\subsection{Time Complexity of the Algorithm}
Consider the setup \eqref{setup} with $n$ observations, $p$ features and $m$ classes. The time complexity for calculating the $GRG$ in the biggest class containing (say) $n_j$ observations is $O(n_j p)$ and the average processing time for sorting the $GRGs$ is $O(n_j\log{n_j})$. If we assume $d$ iterations are taken for the algorithm to converge, then the algorithm has a complexity of $O(d\;n_j^2\;p\log n_j)$ to impute an $n_j\times p$ matrix. We do this for $m$ classes and thus the time complexity for imputing an $n\times p$ matrix is $O(md\;n_j^2\;p\log n_j)$. Now, generally $n_j < n$ whenever $m > 1$, which implies $\log{n_j} < \log{n}$, and $n_j*m \geq n$ since $n_j$ was the biggest class. This gives rise to the inequality $$O(md\;n_j^2\;p\log n_j) < O(d\;n^2\;p\log n).$$
We initially calculate the Mutual Information of each attribute with the class variable, which takes $O(n\;p)$ time along with the imputation of the mean/mode which again takes $O(p)$ time and choosing an optimum $k$ which takes $O(10*n\;p\;r)$ time if we assume $r$ values of $k$ are tested using 10-fold cross-validation. So our total complexity becomes $O(md\;n_j^2\;p\log n_j + np + p + 10npr)$ which can be approximated to $O(md\;n_j^2\;p\log n_j)$ if the value of $n_j$ is large compared to $r$. We note that this time complexity is similar to Grey KNNI (GKNN) and Feature-Weighted Grey KNNI (FWGKNN) but less than the $O(d\;n^2\;p\log n)$ complexity of Iterative KNNI (IKNN) and the Grey-Based Nearest Neighbor (GBNN) algorithm \citep{huang2004grey}.

\section{Simulation Studies}
In this section we explore the performance of our proposed \textbf{Class-weighted Grey KNN (CGKNN)} algorithm in recovering missing entries and improving the classification accuracy, and we report on computational efficiency of the algorithm. We compare our method with 6 other well-established methods which are as follows:
 \begin{itemize}
     \item \textbf{MICE (Multiple Imputation using Chained Equations):} Multiple imputation using Fully Conditional Specification (FCS) of the variables of $X$ is implemented by the MICE algorithm developed by Van Buuren and Oudshoorn \citep{buuren1999flexible}. Each variable has its own imputation model. We use the built-in imputation model for continuous data (predictive mean matching), binary data (logistic regression), unordered categorical data (polytomous logistic regression), and ordered categorical data (proportional odds) based on the type of variable encountered during imputation.
     
    \item \textbf{MissForest:} This is an iterative imputation method based on a random forest developed by Stekhoven and Buhlmann \citep{stekhoven2011missforest}. This non-parametric algorithm can handle both continuous and categorical data, and also give an out-of-bag error estimate to estimate the imputation error.
    
    \item \textbf{Iterative $k$-Nearest Neighbor imputation (IKNN):} This method adopts the Euclidean distance to compute similarities between random instances. Mean imputation is regarded as a preliminary estimate. The $k$ closest objects are selected from the candidate data which contain all instances, except the one that is to be imputed. The complete instances are upgraded after first imputation, and the iteration procedure is repeated until reaching the convergence criterion.
    
    \item \textbf{Mutual Information based $k$-Nearest Neighbor imputation (MI-KNN):} This approach first measures the relevance of each feature in the classification problem similar to the approach described in \eqref{MInotion}, and uses a weighted Euclidean distance to measure the distance between instances, with the mutual information being the weights \citep{garcia2009k}. Mean or mode imputation is used as a preliminary estimate. All imputed instances and all complete instances are considered to be known information for estimating missing values iteratively. The missing values are then imputed based on the weighted mean or mode of the nearest neighbours.
    
    \item \textbf{Grey $k$-Nearest Neighbor imputation (GKNN):} This approach is used to handle heterogeneous data (numerical and categorical data). It uses GRA to measure the relationships between instances and seek out the $k$ nearest neighbours to execute the missing values estimation \citep{zhang2012nearest}. The dataset is divided into several parts based on the class label and imputation is performed on each of them. Mean or mode imputation is used as a preliminary estimate. The missing values are then imputed iteratively, based on mean or mode of the nearest neighbors sorted by Grey distance.
    
    \item \textbf{Feature Weighted Grey $k$-Nearest Neighbor (FWGKNN):} This approach employs Mutual Information (MI) to measure inter-feature relevance in the $X$ matrix. It then uses the GRA to find the distance between instances and weighs them by the inter-feature relevance \citep{pan2015missing}. The difference between FWGKNN and our CGKNN algorithm is that the mutual information is computed between the class variable $Y$ and the features $X_{.j}$ in our algorithm whereas it is $I(X_{.i},X_{.j})$ for the FWGKNN algorithm. Our approach is focused towards classification relevance instead of inter-feature relevance.
 \end{itemize}
 
 \subsection{Performance Measure}
We measure the performance of each algorithm according to the following metrics:
\begin{itemize}
    \item Accuracy of Prediction: The \textbf{root mean square error (RMSE)} is used to evaluate the precision of imputation as follows:
        \begin{equation}
            RMSE = \sqrt{\frac{1}{m} \sum_{i=1}^m {(e_i - \Tilde{e}_i)}^2},
        \end{equation}
    where $e_i$ is the true value, $\Tilde{e}_i$ is the imputed value of the missing data, and $m$ denotes the number of missing values.
    
    \item \textbf{Classification accuracy (CA):} After estimating the missing values, an incomplete dataset can be treated as a complete dataset. We used classification accuracy to evaluate all the imputation methods and to show the impact of imputation on the accuracy of classification as follows:
        \begin{equation}
            CA = \frac{1}{n} \sum_{i=1}^n I (IC_i = RC_i),
        \end{equation}
    where $n$ denotes the number of the instances in the dataset, $IC_i$ and $RC_i$ are the classification results for the $i$-th instance and the true class label, and $I(.)$ is the indicator function.
\end{itemize}

\subsection{Simulation Scenarios:}

\subsubsection{Missing Completely at Random (MCAR) Example}
\begin{figure}[h!]
    \begin{center}
        \includegraphics[width=0.6\textwidth]{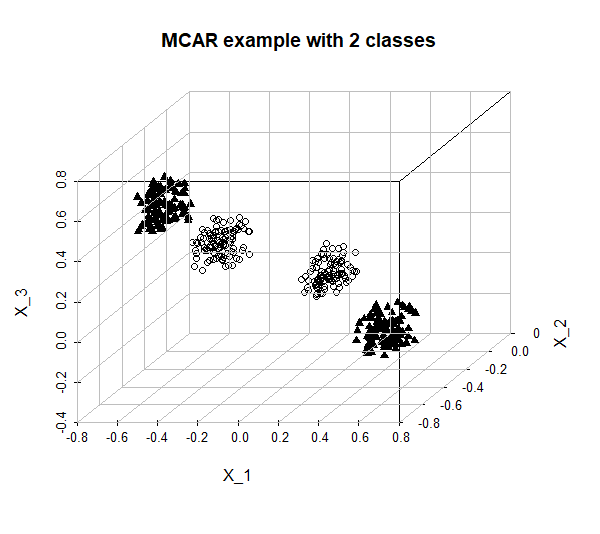}
        \caption{3D centers of each class represented without noise variables}
        \label{fig1}
    \end{center}
\end{figure}

We use an artificial example to demonstrate the effect of mutual information with the class variable while selecting the k-Nearest Neighbors. We took a separable example with four cubes drawn in a three dimensional space. Fig. \ref{fig1} shows this artificial problem. Two cubes belong to class 1, and they are centered on $(0,0,0)$ and $(-0.2,-0.4,0.4)$. The remaining two cubes are labeled with the class 2, being centered on $(-0.6,-0.6,0.5)$ and $(0.4,-0.2,-0.2)$.  In all the cubes, the radius is equal to 0.10, and they are composed of 100 samples which are uniformly distributed inside the cube. In this problem, the MI values between the three attributes and the target class are computed: 0.40 for $x_1$, 0.28 for $x_2$, and 0.21 for $x_3$.

To this 3 dimensional, 2 class dataset we add 20 U[-1,1] variables. For these irrelevant variables, the MI between the feature and class variable is almost 0. We try to find out what happens when we add irrelevant attributes to classification. We insert $10\%$ and $20\%$ of missing data to $x_1$, which is most relevant according to MI. The missingness of data in $x_2$ is generated completely at random, which means it does not depend on the variable values in the matrix $X$.

This advantage is clearer for higher percentages of missing values, as it is shown by the differences in Table 1. The class weighting procedure based on the MI concept discards the irrelevant features, and the selected neighbourhood for missing data estimation tends to provide reliable values for solving the classification task. We provide a detailed analysis of how all the 6 algorithms performed in this simulation setting with $n = 400,\;p = 23$ and $m = 2$ classes in Table 1. Note that we used predictive mean matching as the imputation model for MICE.

\begin{table}[h!] \label{table1}
\centering
\caption{RMSE upon convergence for the toy dataset}
\begin{tabular}{c c c c c c c c}
\hline
Missing Rate & MICE & MissForest & IKNN & MI-KNN & GKNN & FWGKNN & CGKNN  \\
\hline
10\% & 0.2205 & 0.1902 & 0.1807 & 0.1423 & 0.1443 & 0.1355 & \textbf{0.1283} \\
20\% & 0.2607 & 0.2103 & 0.2168 & 0.1852 & 0.1872 & 0.1603 & \textbf{0.1498}\\
\hline
\end{tabular}
\end{table}

We also calculated the classification accuracy using the Naive Bayes method on the non-imputed and imputed datasets with 10\% and 20\% missing data, with the help of 10-fold cross validation process. The resulting improvement in accuracy for both the cases is highest for our CGKNN Algorithm, as shown in Table 2.

\begin{table}[h!] \label{table2}
\centering
\caption{Classification Accuracy (\%) for the toy dataset}
\begin{tabular}{c c c c c c c c}
\hline
Missing Rate & 10\% & 20\% \\
\hline
No Imputation & 83.42 & 76.84 \\
MICE & 85.24 & 79.14 \\
MissForest & 86.75 & 79.10 \\
IKNN & 88.23 & 80.54 \\
MI-KNN & 93.71 & 90.24 \\
GKNN & 92.20 & 80.20 \\
FWGKNN & 94.19 & 92.84\\
CGKNN & \textbf{96.92} & \textbf{94.22}\\
\hline
\end{tabular}
\end{table}

\subsubsection{Missing at Random (MAR) Example}
For this section, we illustrate how our method performs with respect to the six other techniques. We simulate our data from the multivariate normal distribution and then artificially introduce missingness in the data, at random (MAR), by letting the probability of missingness depend on the observed values. We take the number of classes $m = 4$, the number of attributes $p=5$ and generate $n=100$ observations for each class. Specifically,
$$X_{i}^{(k)} \sim {N(\mu^{(k)}, \Sigma^{(k)})}, i = 1, 2,..., 100, k = 1, ..., 4, $$
where $k$ stands for the $k-$th class, $\mu^{(k)} \sim{U[-1,1]^5}\; \forall \; k$ and $\Sigma^{(k)}$'s are randomly generated $5*5$ positive definite matrices using partial correlations \citep{joe2006generating}. This simulation procedure ensures us that we do not have the same mean and variance for two different classes during simulation. Also, the missingness is induced using a logistic model on the missingness matrix $D$. In real life, we often encounter covariates which are demographic in nature and thus non-missing. For this example, we assume $X^{(k)}_{i1}, X^{(k)}_{i2}$ and $X^{(k)}_{i3}$ to be non-missing and the missingness of $X^{(k)}_{i4}$ and $X^{(k)}_{i5}$ to be dependent on these demographic, non-missing variables, for each class $k$. Recall the $n*p$ missing matrix $D$, which we modify to a layered 3D matrix $D^{(k)}, k=1,..,4$ with $n*p*m$ entries. We assume $D^{(k)}_{i1},D^{(k)}_{i2},D^{(k)}_{i3}$ to be all $1$ and 
\begin{eqnarray}
 D^{(k)}_{i4} \sim Ber(logit({p_{11}+p_{21}*X^{(k)}_{i1}+p_{31}*X^{(k)}_{i2}+p_{41}*X^{(k)}_{i3}})),\\
D^{(k)}_{i5} \sim Ber(logit({p_{12}+p_{22}*X^{(k)}_{i1}+p_{32}*X^{(k)}_{i2}+p_{42}*X^{(k)}_{i3}})) 
\end{eqnarray}
where $logit(x) = \frac{\mathrm{e}^x}{1+\mathrm{e}^x}$ and $p_{ij} 's$ are vectors of size $p=5$ chosen by us.

We provide a detailed analysis of how all the 6 algorithms performed in this simulation setting with $n = 100,\;p = 5$ and $m = 4$ classes in Table 3. Note that we used predictive mean matching as the imputation model for MICE.

\begin{table}[h!] \label{table3}
\centering
\caption{RMSE upon convergence for the toy MAR dataset}
\begin{tabular}{c c c c c c c c}
\hline
Missing Rate & MICE & MissForest & IKNN & MI-KNN & GKNN & FWGKNN & CGKNN  \\
\hline
10\% & 0.1201 & 0.1102 & 0.0907 & 0.0871 & 0.0899 & 0.0829 & \textbf{0.0827} \\
20\% & 0.1584 & 0.1677 & 0.1250 & 0.1123 & 0.1208 & 0.1096 & \textbf{0.1075}\\
\hline
\end{tabular}
\end{table}

We also calculated the classification accuracy using the Naive Bayes method on the non-imputed and imputed datasets with 10\% and 20\% missing data, with the help of 10-fold cross validation process. The resulting improvement in accuracy for both the cases is highest for our CGKNN Algorithm, as shown in Table 4.

\begin{table}[h!] \label{table42}
\centering
\caption{Classification Accuracy (\%) for MAR dataset}
\begin{tabular}{c c c c c c c c}
\hline
Missing Rate & 10\% & 20\% \\
\hline
No Imputation & 75.13 & 72.50 \\
MICE & 79.09 & 77.32 \\
MissForest & 83.75 & 80.11 \\
IKNN & 85.48 & 83.91 \\
MI-KNN & 88.26 & 86.37 \\
GKNN & 86.38 & 84.12 \\
FWGKNN & 89.01 & 87.30\\
CGKNN & \textbf{90.29} & \textbf{87.31}\\
\hline
\end{tabular}
\end{table}

\section{Applications to UCI Machine Learning Repository Datasets}
We evaluate the effectiveness of our imputation algorithm on 3 datasets obtained from UCI Machine Learning Repository \citep{newman2008uci}, the Iris (Fisher's Iris Dataset), Voting and Hepatitis datasets, having respectively, characteristics mentioned in Table5.

\begin{table}[h] \label{table5}
\centering
\caption{Characteristics of the UCI Datasets used for data analysis}
\begin{tabular}{c c c c c c}
\hline
Dataset & Instances & Features & Classes & Feature type & \% Missing Rate \\
\hline
Iris & 150 & 4 & 3 & Continuous & 0\\
Voting & 435 & 15 & 2 & Categorical & 4.14\\
Hepatitis & 155 &  19 & 2 & Mixed (both) & 5.39\\
\hline
\end{tabular}
\end{table}

\begin{figure}[H]
    \centering
    \includegraphics[scale = 0.5]{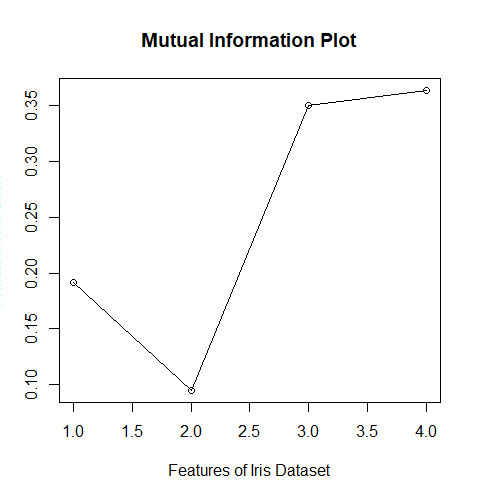}
    \caption{MI for the sepal and petal lengths and widths of Iris Dataset}
    \label{fig4}
\end{figure}

\begin{figure}[H]
    \centering
    \includegraphics[scale = 0.5]{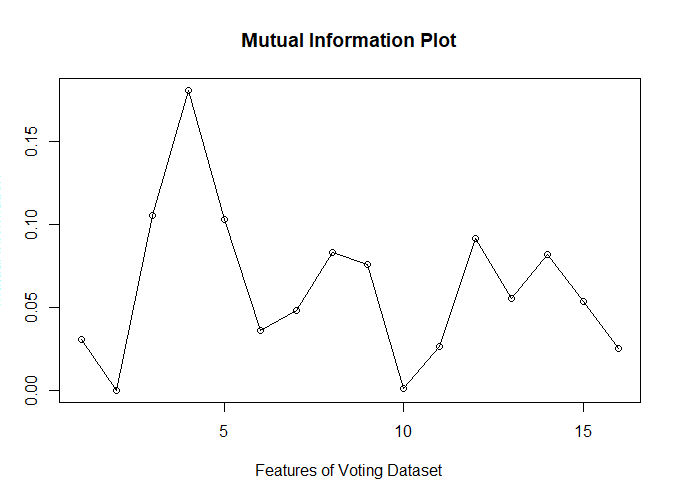}
    \caption{MI for Voting Dataset}
    \label{fig5}
\end{figure}

\begin{figure}[H]
    \centering
    \includegraphics[scale = 0.5]{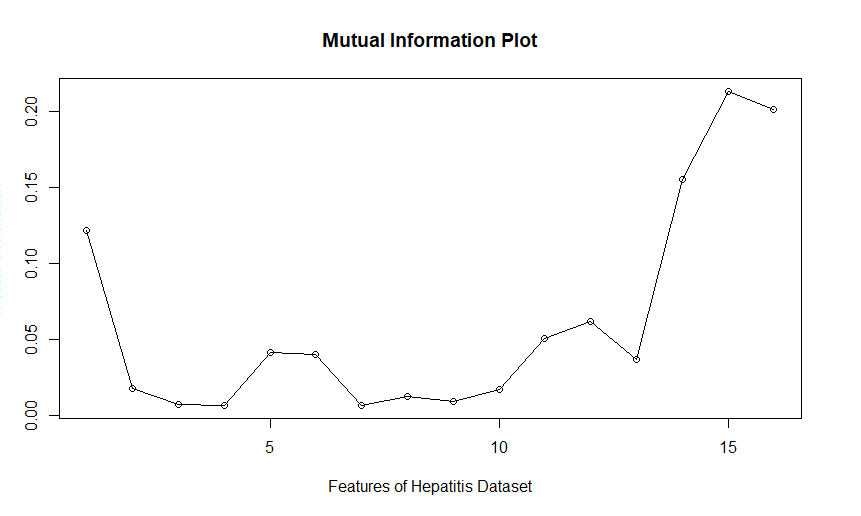}
    \caption{MI for Hepatitis Dataset}
    \label{fig6}
\end{figure}

We represent the Mutual Information (MI) of each feature in these datasets in the graphs shown in Fig. \ref{fig4} -  \ref{fig6}, and use it as the weights for our CGKNN algorithm.

We then introduce 3 different rates of artificial missingness at random (MAR) - 5\%, 10\% and 20\%. Then we run each of the imputation algorithms and calculate the RMSE of imputation after each algorithm converges. For MICE, we used predictive mean matching for continuous variables and polytomous logistic regression for categorical variables. Looking at Table~\ref{table4} - Table 6 note that in almost all cases, our algorithm CGKNN performs better than the other algorithms, usually at higher percentages of missing values. MICE performs the worst in most cases, followed by MissForest, probably because they do not take into account any sort of feature relevance.

\begin{table}[h!] \label{table4}
\centering
\caption{Comparison of RMSE of Iris Dataset}
\begin{tabular}{c c c c c c c c}
\hline
Missing Rate & MICE & MissForest & IKNN & MI-KNN & GKNN & FWGKNN & CGKNN  \\
\hline
5\% & 0.0729 & 0.0619 & 0.0588 & \textbf{0.0503} & 0.0534 & 0.0506 & 0.0509 \\
10\% & 0.1205 & 0.1302 & 0.1107 & 0.1025 & 0.1038 & 0.0995 & \textbf{0.0950} \\
20\% & 0.1427 & 0.1420 & 0.1241 & 0.1146 & 0.1246 & 0.1106 & \textbf{0.1001}\\
\hline
\end{tabular}
\end{table}

\begin{table}[h!] \label{table5}
\centering
\caption{Comparison of RMSE of Voting Dataset}
\begin{tabular}{c c c c c c c c}
\hline
Missing Rate & MICE & MissForest & IKNN & MI-KNN & GKNN & FWGKNN & CGKNN  \\
\hline
5\% & 0.0928 & 0.0919 & 0.0874 & 0.0791 & 0.0820 & \textbf{0.0770} & 0.0779 \\
10\% & 0.1029 & 0.1002 & 0.0949 & 0.0870 & 0.0929 & 0.0868 & \textbf{0.0827} \\
20\% & 0.1521 & 0.1601 & 0.1574 & 0.1446 & 0.1099 & 0.1088 & \textbf{0.1049}\\
\hline
\end{tabular}
\end{table}

\begin{table}[h!] \label{table6}
\centering
\caption{Comparison of RMSE of Hepatitis Dataset}
\begin{tabular}{c c c c c c c c}
\hline
Missing Rate & MICE & MissForest & IKNN & MI-KNN & GKNN & FWGKNN & CGKNN  \\
\hline
5\% & 0.0913 & 0.0890 & 0.0785 & \textbf{0.0711} & 0.0792 & 0.0739 & 0.0714 \\
10\% & 0.1029 & 0.1002 & 0.1107 & 0.0870 & 0.1038 & 0.0994 & \textbf{0.0921} \\
20\% & 0.1967 & 0.1858 & 0.1592 & 0.0839 & 0.0980 & 0.0898 & \textbf{0.0823}\\
\hline
\end{tabular}
\end{table}

\begin{figure}[H]
    \centering
    \includegraphics[scale = 0.7]{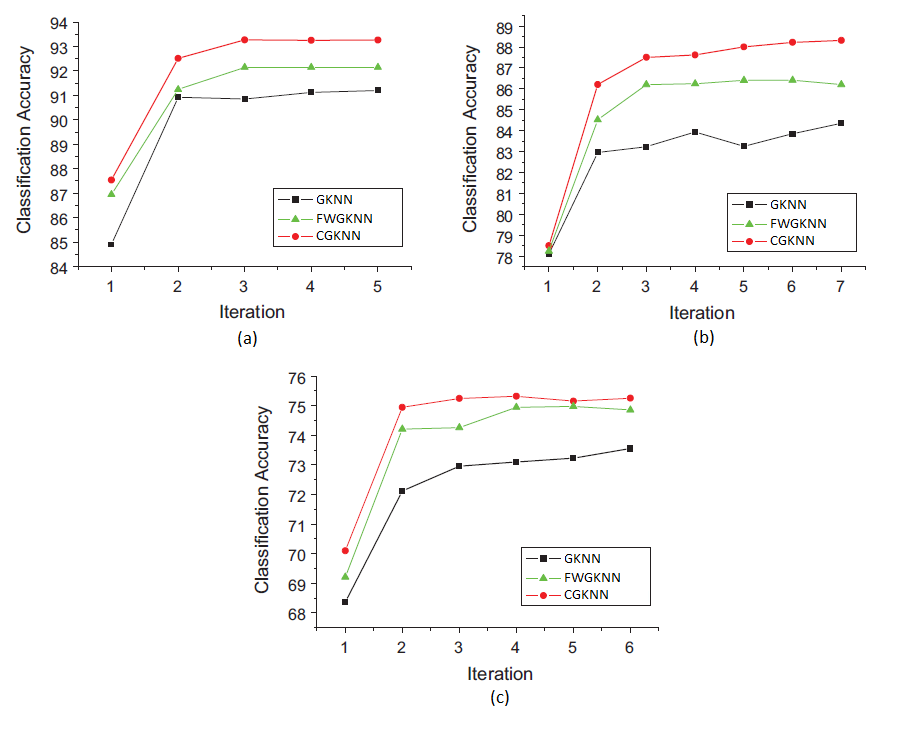}
    \caption{Classification Accuracy for the Iris dataset at (a) 5\% (b) 10 \% and (c) 20 \% rates of missingness after using an imputation algorithm}
    \label{fig7}
\end{figure}

We use a Naive Bayes classifier on the Iris dataset with 5\% - 20\% missingness and see that our CGKNN algorithm outperforms the closest approach FWGKNN and also GKNN, when used as an imputation approach before the classifier. The CGKNN algorithm also converges quite fast with respect to classification accuracy as shown in Fig.~\ref{fig7}.

\section{Discussion}
Missing data is a classical drawback for most classification algorithms. However, most of the missing data imputation techniques have been developed without taking into account the class information, which is always present for a supervised machine learning problem. $k$-Nearest Neighbors is a good technique for imputation of missing data and has shown to perform well against many other imputation procedures. We have proposed a method which not only takes into account the class information, but also uses a better metric to calculate the nearest neighbors in KNN imputation. Our Class-weighted Grey $k$-Nearest Neighbor (CGKNN) approach has same time complexity as the previous algorithms and even better than some KNN imputation algorithms like Grey-Based k-Nearest Neighbor (GBNN) and Iterative $k$-Nearest Neighbor (IKNN) imputation. We have shown that it outperforms all the other algorithms in simulated settings, as well as high rates of missingness in actual (non-simulated) datasets as far as imputation is concerned. We also show that it improves the accuracy of classification better than other imputation procedures. We do not make any assumptions regarding the variables of the feature matrix and thus, for any classification problem, our method can be used to impute missing data in the feature matrix.

However, an open problem is the selection of $k$ in our nearest neighbors approach and we have chosen it through cross-validation and this method takes time. The reason why $k$ is difficult to predict is because we do not have anything to validate the true value of $k$ in our datasets. A potential future research could be to select the value of $k$ in a smart, effective manner without involving cross-validation. Our algorithm has not been theoretically proven to converge, although it has been shown empirically. Finding the rate of convergence of our CGKNN algorithm is a good theoretical problem to consider.

Another potentially interesting future research topic would be to extend this idea to regression problem where the outcome $Y$ is continuous instead of categorical. The imputation of the data matrix $X$ could be done with the help of information from $Y$ since they are assumed to be related in a regression setting. We could also look into better methods of measuring the relationship between the features and class variable than mutual information (MI) and then use them as weights for the Grey distance. Another potential future research paper is to develop an algorithm which imputes and classifies simultaneously, thus yielding a better classification in a single step instead of imputation and classification at two different stages. This idea has already been worked on in Learning Vector Quantization (LVQ) \citep{villmann2006comparison} but can be vastly improved.

The most difficult challenge, however, to find imputation techniques when the data is Not Missing at Random (NMAR). It is difficult to model this setting without making strong assumptions, and much development is still possible in that area. The main difficulty is to tackle the problem without assuming anything that may cause a bias - and that is not possible. Hopefully, new ideas will crop up in the future which will make NMAR problem easier to handle.

\section{Acknowledgements}
We would like to acknowledge support for this project from the National Cancer Institute grant P01 CA142538.

\vskip 0.2in 

\bibliography{Referencias}

\appendix



\end{document}